\documentclass{article}
\usepackage{spconf,amsmath,graphicx}
\usepackage{booktabs}
\usepackage{amssymb}
\usepackage{hyperref}

\let\oldbibliography\thebibliography
\renewcommand{\thebibliography}[1]{%
  \oldbibliography{#1}%
  \setlength{\itemsep}{3pt}%
}

\title{Invariances and Data Augmentation for Supervised Music Transcription}
\name{John Thickstun$^{\star}$ \qquad Zaid Harchaoui$^{\star}$ \qquad Dean Foster$^{\dagger}$ \qquad Sham M. Kakade$^{\star}$}
\address{$^{\star}$ University of Washington, $^{\dagger}$ Amazon  \\
\texttt{\{thickstn,sham\}@cs.washington.edu},~~\texttt{zaid@uw.edu},~~\texttt{dean@foster.net} \\}
\begin{document}
%
\maketitle
\begin{abstract}
This paper explores a variety of models for frame-based music transcription, with an emphasis on the methods needed to reach state-of-the-art on human recordings. The translation-invariant network discussed in this paper, which combines a traditional filterbank with a convolutional neural network, was the top-performing model in the 2017 MIREX Multiple Fundamental Frequency Estimation evaluation. This class of models shares parameters in the log-frequency domain, which exploits the frequency invariance of music to reduce the number of model parameters and avoid overfitting to the training data. All models in this paper were trained with supervision by labeled data from the MusicNet dataset, augmented by random label-preserving pitch-shift transformations.
\end{abstract}
\section{Introduction}
\label{sec:intro}

The prominent success of deep learning in vision has popularized end-to-end learning approaches to supervised classification tasks. These methods depend upon large quantities of labeled training data. Many researchers have proposed supervised approaches to music transcription using synthesized training data, including recent MIREX participants \cite{troxel2016,marolt2016,mita2017}. While synthesized recordings provide an effectively infinite supply of labeled data, MIREX results suggest that models trained on synthetic data do not generalize well to human recordings.

There is a large and growing collection of human recordings annotated with labels suitable for supervision of music transcription \cite{goto2003, emiya2010, raffel2016, thickstun2017}. This prompts us to investigate models that can make effective use of this data. While the amount of available data is substantial, it is not infinite. Furthermore, frames of music sampled from an audio recording are more highly correlated than, for example, a pair of images from ImageNet. We therefore focus our attention on models and data augmentation techniques that incorporate prior knowledge of invariances in the problem domain to efficiently use the data.

Recent work shows that end-to-end architectures construct a first-layer feature representation that is qualitatively comparable to classical frequency filterbank transforms such as the STFT or CQT \cite{dieleman2014,thickstun2017}. We will see that deep end-to-end models overfit to current datasets of human performances. This leads us to reconsider filterbanks as a low-level representation of musical audio. By replacing the first layer of an end-to-end network with a fixed filterbank transform, we dramatically reduce the number of model parameters and the attendant risk of overfitting.

The filterbank representation has a further advantage over end-to-end learning: its channels are ordered from low to high frequency. This order structure introduces a topology on the channel axis (the frequency domain) that motivates a convolutional architecture, analogous to how the Euclidean structure of $\mathbb{R}^2$ motivates the classic ConvNets used in computer vision. Constructing a ConvNet on the channels of a filterbank representation yields performance gains from parameter sharing that are not obviously replicable in an end-to-end architecture.

In this paper, we explore convolutional architectures that exploit the topological structure of a frequency filterbank representation. In Section \ref{sec:rel} we discuss related work on music transcription. In Section \ref{sec:arch} we will describe the network architectures of the models under consideration. In Section \ref{sec:opt} we discuss optimization of these networks, including label-preserving transformations that augment the size of the training data. We present our qualitative results in Section \ref{sec:res} along with quantitative results on the MusicNet dataset and MIREX evaluation dataset.

\section{Related Works}
\label{sec:rel}

To the best of our knowledge, music transcription was first considered as a supervised learning problem in \cite{poliner2007} using labels obtained from MIDI files to train an SVM on the spectrograms of synthesized recordings of these MIDIs. Subsequent work on supervised transcription improves upon these results in two directions: use of more sophisticated models, and construction of datasets of recorded human performances (as opposed to synthesized data).

Labels on music recordings are typically obtained in one of two ways. One approach is to perform music on instruments that are wired to record a MIDI transcription as they are played. These transcriptions are precise time-aligned labels for a performance. The commercially available Yamaha Disklavier piano is wired in this way, and has lead to the creation of datasets such as MAPS \cite{emiya2010}. The second approach is to solve an alignment problem, warping a musical score onto a given recording. This alignment can be constructed using an optimal-alignment protocol, as in the SyncRWC \cite{goto2003}, Lakh \cite{raffel2016}, and MusicNet \cite{thickstun2017} datasets. Or it can be constructed using information supplied by a human annotator, for example the Su dataset used for the MIREX evaluation \cite{su2015}.

The development of models for music transcription conceptually factors into two subproblems: acoustic modeling and time series prediction. In this paper, we focus on the acoustic modeling problem, as introduced in \cite{poliner2007}. Recent developments in this area model the acoustics using deep neural networks \cite{nam2011,trabelsi2017} or convolutional neural networks \cite{bittner2017, pons2017, thickstun2017}. Some recent work explores hybrid models that combine a deep or convolutional acoustic model with a recurrent time-series model to jointly estimate transcriptions \cite{sigtia2015,sigtia2016}.

Choosing an appropriate model for a supervised learning problem requires consideration of both the structure of the problem and the available data. A highly biased model can compensate for a smaller dataset at the risk of making overly powerful assumptions about the problem structure; a more general model requires more data to overcome variance.  The frequency-invariance ideas that lead to the best models presented in this paper are anticipated in \cite{sigtia2016,bittner2017,pons2017}. Our contribution is to demonstrate that this class of models represents a good bias-variance tradeoff for current datasets. Our dataset augmentation techniques are inspired by analogous transformations introduced by the vision community for image classification \cite{simard2003,krizhevsky2012} and extensions of these ideas to audio \cite{mcfee2015}. 

\section{Methods}
\label{sec:arch}

Given an audio segment $\textbf{x} \in \mathcal{X}$, we seek to predict the notes present at the midpoint of $\textbf{x}$, which we encode as a binary label vector $\textbf{y} \in \{0,1\}^{128}$. We accomplish this task by learning a feature map $f_\theta : \mathcal{X} \to \mathcal{H}$, along with a multivariate linear regression to estimate $\mathbf{\hat y}$ given the learned representation $f(\textbf{x})$. We consider variants of four network architectures $f_\theta$ for this purpose. 

We take $\textbf{x}$ to be real-valued audio frame with values in the range $[-1,1]$, sampled at $44.1$kHz. We preprocess $\textbf{x}$ by the normalization $\textbf{x} \mapsto \textbf{x}/\|\textbf{x}\|_2$; this can be interpreted physically as normalizing the audible volume of each frame. The first layer of every network considered in this paper is a strided convolution with a $4,096$-sample receptive field and a $512$-sample stride. We use a frame of $16,384$ samples, resulting in $25 = (16384-4096)/512$ regions per frame. 

The $16,384$-sample frame size reflects a tradeoff between a shorter frame, which could miss important context for the classification task, and a longer frame, which has diminishing returns at computational cost. Very long frames grow the number of parameters in the model to the point of overfitting. The $512$-sample stride is subject to a similar tradeoff.

\subsection{Two layer networks}

The simplest model we consider is a two layer network. For each region of the layer-one convolution we compute a filterbank representation of the input, creating a spectrogram representation $\mathcal{H}$ at layer two. We perform linear classification on $\log \mathcal{H}$, the pointwise logarithm of the spectrogram. We consider several variants on the choice of filterbank below, as well as an end-to-end architecture where the filters are learned from data.

\textbf{(Short-time Fourier transform)} This is the classical filterbank consisting of Fourier coefficient magnitudes. We truncate the magnitude spectrum at $6$kHz because we find that frequencies above this cutoff do not meaningfully improve classification accuracy.

\textbf{(Log-spaced filterbank)} This filterbank consists of $512$ sine and cosine filters with logarithmically-spaced frequencies ranging from $50$Hz to $6$kHz. For each filter pair $\textbf{w}_\text{k,sin},\textbf{w}_\text{k,cos}$, we compute inner products with the input region $\textbf{x}_t \in [-1,1]^{4096}$ and sum the square of these values, analogous to the STFT:
\begin{equation*}
\text{filter}_k = (\textbf{w}_\text{k,sin}^T\textbf{x}_t)^2 + (\textbf{w}_{k,cos}^T\textbf{x}_t)^2.
\end{equation*}

\textbf{(Windowed filterbank)} Here we apply the cosine window $1 - \cos(t)$ to each filter in our filterbank. This combats the spectral leakage phenomenon caused by boundary effects introduced by the finite-window frequency analysis \cite{rabiner2007}. We will examine the effects of windowing on both the STFT and log-spaced filterbank.

\textbf{(Learned filterbank)} In this architecture, the filter coefficients $\textbf{w}_k$ are learned as parameters in the classifier optimization. This network is discussed at length in \cite{thickstun2017}. We will revisit these results in Section \ref{sec:res} and compare them to the hand-crafted filterbanks discussed above.

\subsection{Three layer networks}

A natural extension of the two layer networks discussed above is a three layer network with a fully connected layer interposed between the layer-one convolutions and the linear output layer. If we interpret the output of layer one as a spectrogram, then this intermediate layer captures non-linear relationships between features of this spectrogram. A filter in layer two might be sensitive to a particular chord, for example, or to a certain progression of notes. In Section \ref{sec:res} we report results for two three-layer networks, one trained with a fixed log-spaced, cosine-windowed filterbank at layer one, and the other trained end-to-end from the raw audio.

In this vein, it is possible to build much deeper models on top of either the raw audio or filterbank representation. These ideas are explored in \cite{trabelsi2017}. However, as we will see in Section \ref{sec:res}, simply building a deeper architecture does not appear to boost performance for the note classification task. Instead, we will turn to translation-invariant networks that introduce additional layers to capture specific invariances in the data.

\subsection{Translation-invariant networks}

\begin{figure}[htb]
\label{fig:arch}
\includegraphics[scale=.37]{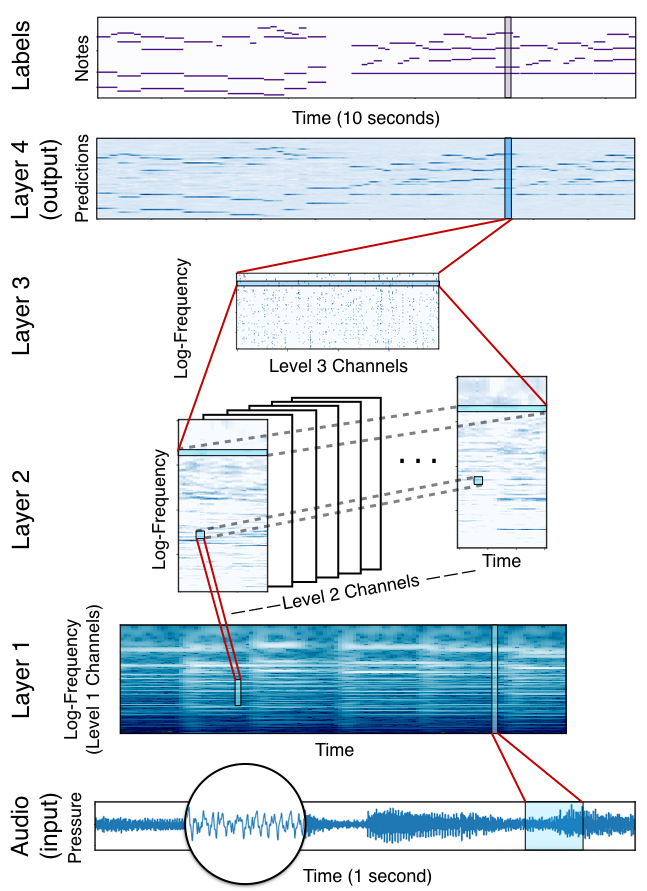}
\caption{A translation-invariant network for note classification. Audio input maps to Layer 1 according to the log-spaced, cosine-windowed filterbank described in Section \ref{sec:arch}.1. Layer~1 maps to Layer 2 by convolving a set of $128\times 1$ learned filters along the log-frequency axis at each fixed time location. Layer 2 maps to Layer 3 by convolving again along the log-frequency axis, this time with a set of filters of height 1 that fully connect along the time and channel axes of Layer 2. Notes are predicted at Layer 4 by linear classification on the Layer 3 representation.}
\end{figure}

The translation-invariant network is built on top of a filterbank, with two learned representational layers. See Figure~1 for a visual schematic and description of this architecture. A handcrafted layer-one filterbank is crucial to support the translation-invariant filters at layer two. Because the filters are frequency-ordered, the layer-two filters can learn patterns that are invariant to translations in frequency. Consider, for example, a major triad chord. This pattern is preserved under linear translations in log-frequency space. In the translation-invariant architecture, a single filter consisting of only $128$ parameters could be sensitive major triads rooted at arbitrary frequencies. Compare this situation to a fully connected three layer network (Section \ref{sec:arch}.2) which would require a separate filter to identify this chordal pattern at each location in the log-frequency spectrum.

The preceding arguments about music-theoretic concepts like intervals and triads are contingent on the use of a log-frequency filterbank for layer one. If we used a linear filterbank (for example, the STFT) then a linear shift in the frequency domain would correspond to a non-linear shift in the musical relationships between notes (low notes would translate further than high notes) due to the human ear's logarithmic perception of frequency. On the other hand, the physics of audio (for example, overtones) exist on a linear scale. A model that uses a linear filterbank can exploit translation invariances in the physics. We find empirically that log-scale invariance yields greater performance gains for note classification than linear-scale invariance.

\subsection{Channel convolutions}

Finally, we consider an end-to-end network that uses the same architecture as the translation-invariant network, but treats the weights in the layer-one filterbank as optimization parameters (i.e. the filterbank is learned). Because the weights in layer one are learned from a random initialization, it is not clear that the learned filters will be ordered by frequency or that their output channels will exhibit any topological structure. However, because the layer-two convolutions in this architecture are designed to exploit local topological structure in the channels, we might hope that end-to-end parameter optimization would find a good topological structure for this space, a kind of self-organizing map \cite{kohonen1990}.

\section{Training}
\label{sec:opt}

We trained our models on the MusicNet dataset \cite{thickstun2017} with minibatch stochastic gradient optimization ($150$ samples per minibatch) using momentum ($\rho = .95$). The models are implemented in TensorFlow on an NVIDIA 1080Ti GPU. The final network weights are computed from a moving average of iterates with a decay factor of $2\times 10^{-4}$.

\textbf{Data augmentation.} We augment our data by stretching or shrinking our input audio with linear interpolation. This corresponds to a pitch-shift in the frequency domain. For small shifts ($\pm 5$ semitones or less) the transformed audio sounds natural to the human ear. Randomly shifting each data points in a minibatch by an integral number of semitones in the range $[-5,5]$ augments the dataset by an order of magnitude. And the translational nature of this augmentation reinforces the architectural structure of the translation-invariant networks described in Section~\ref{sec:arch}. In addition to an integral semitone shift, we also apply a continuous shift to each data point in the range $[-.1,.1]$. This makes the models more robust to tuning variation between recordings.

\section{Results}
\label{sec:res}

Our best translation invariant network achieves $77.3\%$ average precision on MusicNet, outperforming the previous state of the art reported in \cite{trabelsi2017}, and popular commercial software \cite{melodyne}. Furthermore, we find that the translation-invariant architecture substantially outperforms previously proposed end-to-end models on this dataset. 

\begin{table}[h]
 \centering
  \begin{tabular}{lccc}
   \toprule
    \cmidrule{1-4}
    Model & Avg. Prec. & Acc. & Err. \\
    \midrule
    \textbf{filterbanks}\\
    STFT (no compress) & 40.4 & 15.9 & .860 \\
    STFT & 60.4 & 36.2 & .681 \\
    log frequencies & 62.7 & 39.8 & .646 \\
    cosine windows & 66.1 & 38.7 & .637 \\
    log + windows & 66.7 & 38.9 & .633 \\
    three layer network & 73.8 & 51.4 & .541 \\
    \hline
    \textbf{end-to-end}\\
    learned filterbank \cite{thickstun2017} & 67.8 & 48.9 & .634 \\
    three layer network & 70.8 & 48.8 & .558\\
    deep complex \cite{trabelsi2017} & 72.9 & - & - \\
    channel convolution & 73.3 & 50.4 & .531 \\
    \hline
    \textbf{translation-invariant}\\
    baseline & 76.5 & 53.2 & .496 \\
    pitch-shift & 77.1 & 54.5 & .482 \\
    wide layer 3 & \textbf{77.3} & \textbf{55.3} & \textbf{.474} \\
    \hline
    \textbf{commercial software}\\
    Melodyne \cite{melodyne} & 58.8 & 41.0 & .760 \\
   \bottomrule
 \end{tabular}
 \caption{Average Precision, Accuracy, and Error for each of the models discussed in this paper, evaluated using the test set from \cite{thickstun2017}. Average Precision is computed by scikit-learn \cite{scikit-learn}; Accuracy and Error use mir\_eval \cite{raffel2014}. The Accuracy and Error scores are assume a global prediction threshold of~0.4.}
\end{table}

Performance is highly sensitive to layer one. While a naive filterbank performs poorly, log-spaced, cosine-windowed filters approach the performance of a learned filterbank (see Table 1; compare ``log + windows'' to ``learned filterbank''). Also, note the importance of the compressive non-linearity used for all models except the one marked ``no compress.''

We consider three variants of the translation invariant architecture: a baseline, the same model optimized with pitch-shift transformations, and a model with a large number of hidden nodes at layer three ($4,096$ versus $256$). Pitch-shifting is a clear improvement over the baseline. Increasing the number of nodes at level 3 is beneficial but requires dramatically more nodes for small performance gains; there may be an opportunity to model the time-domain structure captured in this layer more efficiently. We do not observe performance increases by adding nodes to layer two (we use just $128$ layer-two filters) suggesting that the translation-invariant architecture efficiently captures frequency-domain structure.

The end-to-end architectures significantly underperform the corresponding architectures with a handcrafted layer-one filterbank. Because filterbanks are essentially realizable in an end-to-end architecture (see \cite{thickstun2017}; Section 4.3) we infer that these optimizations have converged to either a local minimum or a saddle point. In particular, the learned layer-one weights of the channel convolution model exhibit some structure between neighboring filters, but not the global frequency ordering exhibited by the hand-crafted filterbanks.

Regarding the MIREX 2017 evaluation results, we remark that MHMTM1 \cite{mita2017} is a neural network trained on synthesized data. Several such models appeared in recent years at MIREX \cite{troxel2016,marolt2016,mita2017}. Because these networks can be fed an effectively infinite stream of training data, the efficiencies considered in this paper are not relevant and a wide variety of network architectures could fit well to the training data. The fact that these networks do not generalize well to benchmark data underscores the importance of training on datasets of human performances.

\begin{table}[h]
 \centering
  \begin{tabular}{lcccc}
   \toprule
    \cmidrule{1-5}
    Model & Prec. & Rec. & Acc. & Etot \\
    \midrule
    \textbf{MIREX 2009 Dataset}\\
    THK1 & 82.2 & 78.9 &  \textbf{72.0} & \textbf{.316} \\
    KD1 & 72.4 & 81.1 & 66.9 & .419 \\
    MHMTM1 & 72.7 & 78.2 & 65.5 & .441 \\
    WCS1 & 64.0 & 80.6 & 59.3 & .569 \\
    ZCY2 & 62.7 & 56.2 & 50.6 & .601 \\
    \hline
    \textbf{Su Dataset}\\
    THK1 & 70.1 & 54.6 & \textbf{51.0} & \textbf{.529} \\
    KD1 & 45.9 & 45.0 & 38.1 & .745 \\
    WCS1 &  63.6 & 39.7 & 35.7 & .700 \\
    MHMTM1 & 61.2 & 36.8 & 35.2 & .676 \\
    ZCY2 & 40.9 & 28.2 & 26.2 & .799 \\
   \bottomrule
 \end{tabular}
 \caption{MIREX 2017 results for the top 5 participants by accuracy in each category of the Multiple Fundamental Frequency Estimation challenge. THK1 is the wide layer 3 translation-invariant model described in this paper.}
\end{table}

\textbf{Conclusion.} Regarding models: the best-performing translation invariant model naively boosts the number of features at level three to integrate temporal information. If we could construct an invariance (perhaps a scale- or elastic-invariance) in this layer, it might boost performance like the translation invariance at layer one. Regarding dataset augmentation, pitch-invariance is one of many possible label-preserving transformations. Effective noise injection remains an open problem. The authors tried adding Gaussian white noise and saw no performance gains. But we believe that a more realistic noise model could have a strong impact on transcription performance.

\textbf{Acknowledgements.} This work was supported by NSF Grant DGE-1256082, the Washington Research Foundation for innovation in Data-intensive Discovery, and the CIFAR program ``Learning in Machines and Brains.''

\bibliographystyle{IEEEbib}
\bibliography{refs}

\appendix

\section{Extended Statistics}

In this appendix, we reiterate our main results (Table 1) on an expanded test set. We observe that multiple data points taken from within one recording are highly correlated. Therefore, to get a representative sample of MusicNet, it is important to hold out a larger test set of recordings than the one introduced in  \cite{thickstun2017}. For the results in Table 3, we compute statistics on the following extended test set (the first three recordings here are the ones from \cite{thickstun2017}):

\begin{itemize}
\item Bach's Prelude in D major for Solo Piano. WTK Book 1, No 5. Performed by Kimiko Ishizaka. MusicNet recording id 2303.
\item Mozart's Serenade in E-flat major. K375, Movement 4 - Menuetto. Performed by the Soni Ventorum Wind Quintet. MusicNet recording id 1819.
\item Beethoven's String Quartet No. 13 in B-flat major. Opus 130, Movement 2 - Presto. Released by the European Archive. MusicNet recording id 2382.
\item Bach's Cello Suite No. 4 in E-flat major, Movement 6 - Gigue. Released by the European Archive. MusicNet recording id 2298.
\item Bach's Violin Partita No. 3 in E major, Movement 6 - Bourree. Performed by Oliver Colbentston. MusicNet recording id 2191.
\item Beethoven's Piano Sonata No. 30 in E major, Op. 109, Movement 2 - Prestissimo. Performed by Paul Pitman. MusicNet recording id 2556.
\item Beethoven's Wind Sextet in E-flat major, Op. 71, Movement 3 - Menuetto - Quasi Allegretto. Performed by the Skidmore Wind Ensemble. MusicNet recording id 2416.
\item Beethoven's Violin Sonata No. 10 in G major, Op. 96, Movement 3 - Scherzo: Allegro - Trio. Performed by the Irrera Brothers. MusicNet recording id 2628.
\item Schubert's Piano Sonata in C minor, D958, Movement 3 - Menuetto Allegro. Released by the Museopen organization. MusicNet recording id 1759.
\item Haydn's String Quartet in D major, Op. 645, Movement 3 - Menuetto Allegretto. Released by the Museopen organization. MusicNet recording id 2106.
\end{itemize}

\begin{table}[h]
 \centering
  \begin{tabular}{lccc}
   \toprule
    \cmidrule{1-4}
    Model & Avg. Prec. & Acc. & Err. \\
    \midrule
    \textbf{filterbanks}\\
    STFT (no compress) & 40.5 & 17.4 & .855 \\
    STFT & 62.9 & 42.4 & .634 \\
    log frequencies & 64.3 & 44.5 & .619 \\
    cosine windows & 66.1 & 44.1 & .618 \\
    log + windows & 66.4 & 44.4 & .618 \\
    three layer network & 76.3 & 55.4 & .492 \\
    \hline
    \textbf{end-to-end}\\
    three layer network & 72.9 & 53.0 & .503 \\
    channel convolution & 74.6 & 54.0 & .483 \\
    \hline
    \textbf{translation-invariant}\\
    baseline & 77.8 & 57.7 & .449 \\
    pitch-shift & 79.5 & 59.2 & .432 \\
    wide layer 3 & \textbf{79.9} & \textbf{59.9} & \textbf{.423} \\
    \hline
    \textbf{commercial software}\\
    Melodyne \cite{melodyne} & 57.9 & 39.5 & .744 \\
   \bottomrule
 \end{tabular}
 \caption{Average Precision, Accuracy, and Error for each of the models discussed in this paper, evaluated using the extended test set described in this appendix. Average Precision is computed by scikit-learn version 0.19.1 \cite{scikit-learn} (please note that older versions of scikit-learn contained a bug in the average precision metric implementation; see the release notes for version 0.19.1; all average precision numbers in this paper are computed using the implementation in version 0.19.1). Accuracy and Error use mir\_eval \cite{raffel2014}. The Accuracy and Error scores are assume a global prediction threshold of~0.4.}
\end{table}

\section{Code}

Code for all the experiments presented in this paper is available online at \url{https://github.com/jthickstun/thickstun2018invariances/}. The preprocessed version of MusicNet used in these experiments is available at \url{http://homes.cs.washington.edu/~thickstn/icassp_data.tar.gz} (11Gb download) and pre-trained weights for each model are available at \url{http://homes.cs.washington.edu/~thickstn/icassp_weights.tar.gz} (1.1Gb download). Further instructions on using this code and data can be found in the git repository's README.md document.

\end{document}